%% file: main.tex
\documentclass[runningheads]{llncs}
\usepackage[T1]{fontenc}

\usepackage{caption}
\usepackage{subcaption}

\usepackage{amsmath}
\usepackage{graphicx}
\usepackage{xcolor}
\usepackage{adjustbox}
\usepackage{amssymb}
\usepackage{bbm}
\usepackage{hyperref}

\definecolor{input}{HTML}{9673A6}
\definecolor{output}{HTML}{D79B00}

\DeclareMathOperator*{\argmax}{arg\,max}

\newcommand{\MeanRF}{$F_{mean}$ }
\newcommand{\MaxRF}{$F_{max}$ }
\newcommand{\MeanMaxRF}{$F_{mean\rightarrow max}$ }

\newcommand{\mytitle}{Exploiting Text-Image Latent Spaces for the Description of Visual Concepts}
\title{\mytitle}

\usepackage{filecontents}

\usepackage{booktabs}

\usepackage[normalem]{ulem}
\newcommand{\stkout}[1]{\ifmmode\text{\sout{\ensuremath{#1}}}\else\sout{#1}\fi}

\begin{document}
\author{
    Laines Schmalwasser\inst{1,2}\orcidID{0009-0006-1120-1299} \and
    Jakob Gawlikowski \inst{1}\orcidID{0000-0003-2492-4358} \and
    Joachim Denzler\inst{2}\orcidID{0000-0002-3193-3300} \and
    Julia Niebling \inst{1}\orcidID{0000-0001-5413-2234}
}

\authorrunning{L. Schmalwasser et al.}
%
\institute{
    Institute of Data Science, German Aerospace Center, 07745 Jena, Germany 
    \and
    Computer Vision Group, Friedrich Schiller University Jena, 07743 Jena, Germany
    \email{laines.schmalwasser@dlr.de}
}

\maketitle

\input{sec/abstract}

\input{sec/introduction}
\input{sec/related_work}
\input{sec/method}

\input{sec/experiments}

\input{sec/discussion}
\input{sec/conclusion}

\input{sec/acknowledgements}

\bibliographystyle{splncs04}
\bibliography{main.bib}

\input{sec/appendix}

\end{document}

%% file: sec/abstract.tex
\begin{abstract}
Concept Activation Vectors (CAVs) offer insights into neural network decision-making by linking human friendly concepts to the model's internal feature extraction process.
However, when a new set of CAVs is discovered, they must still be translated into a human understandable description.
For image-based neural networks, this is typically done by visualizing the most relevant images of a CAV, while the determination of the concept is left to humans.
In this work, we introduce an approach to aid the interpretation of newly discovered concept sets by suggesting textual descriptions for each CAV.
This is done by mapping the most relevant images representing a CAV into a text-image embedding where a joint description of these relevant images can be computed. 
We propose utilizing the most relevant receptive fields instead of full images encoded.
We demonstrate the capabilities of this approach in multiple experiments with and without given CAV labels, showing that the proposed approach provides accurate descriptions for the CAVs and reduces the challenge of concept interpretation.
\keywords{XAI, Explainability, Concepts, Textual Description, Text-Image-Embeddings}
\end{abstract}

%% file: sec/introduction.tex
\section{Introduction}

\begin{figure*}[!htb]
    \centering
    \begin{subfigure}[t]{1\textwidth}
        \centering
        \includegraphics[width=\textwidth]{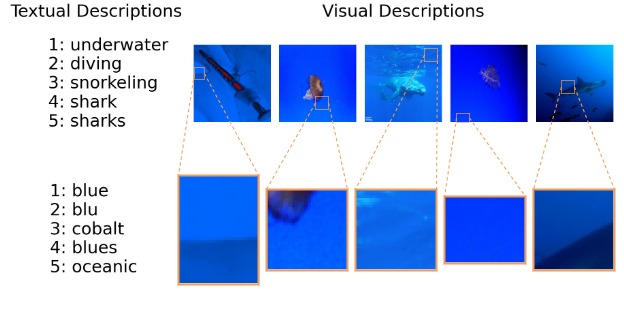}
        \caption{Top derived descriptions: \textit{underwater} vs. \textit{blue}}
        \label{fig:concept_issues_left}
    \end{subfigure}
    \hfill
    \vspace{0.4cm}
    \begin{subfigure}[t]{1\textwidth}
        \centering
        \includegraphics[width=\textwidth]
        {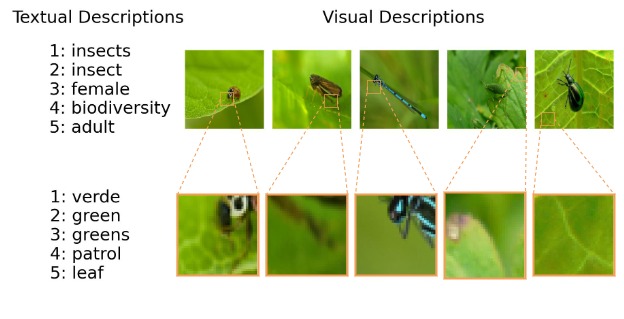}
        \caption{Top derived descriptions: \textit{insects} vs. \textit{verde}
        }
        \label{fig:concept_issues_right}
    \end{subfigure}
\caption{
Examples of two CAVs computed from the first residual block of a ResNet50, trained on Animals with Attributes 2 \cite{xian2018zero}. The first row of each sub\-fig\-ure shows the full representative images of the CAVs and the textual descriptions generated based on the full images. The second row shows the rep\-re\-sen\-ta\-tive receptive fields for the same CAVs and the textual descriptions are derived from the receptive fields.}\label{fig:concept_issues}
\end{figure*}

One major challenge of deep neural networks is their black-box nature which makes the interpretation of their behavior difficult. 
To mitigate this drawback, multiple approaches have been proposed to highlight relevant parts of the input data for a given prediction, for example, LIME \cite{ribeiroWhyShouldTrust2016}, SHAP \cite{lundbergUnifiedApproachInterpreting2017}, GradCAM \cite{selvarajuGradCAMWhyDid2017}, LRP \cite{bachPixelWiseExplanationsNonLinear2015a} and Feature Visualization \cite{olah2017feature}. 
Another idea is to explain the internal mechanism of a deep neural network in terms of concepts that are understandable and easy to communicate to humans \cite{chenConceptWhiteningInterpretable2020,kimInterpretabilityFeatureAttribution2018,reimersDeterminingRelevanceFeatures2020}. 
One attempt to identify such concepts is with so-called Concept Activation Vectors (CAVs) \cite{kimInterpretabilityFeatureAttribution2018}.
A CAV is a vector in the feature space of the activations of a specific network layer. It is designed to point to the direction of activations that are connected to a specific human understandable concept.

The idea behind CAVs is that a human defined concept that contributes to the model decisions has a representation in the model's embedding space. 
For example, the concept \textit{stripe pattern} should have a corresponding representation when the model uses it in the decision-making process to predict a zebra.

In the literature, approaches have been suggested to find CAVs in a supervised and an unsupervised manner:
While for the supervised approaches example images that contain the desired concepts are utilized  \cite{kimInterpretabilityFeatureAttribution2018,moayeriTextToConceptBackCrossModel,yuksekgonulPosthocConceptBottleneck2023}, the unsupervised approaches use, for example, network bottlenecks to extract CAVs \cite{yeh2020completeness,yuksekgonulPosthocConceptBottleneck2023}.

We aim to describe the utilized concepts of a pretrained network without any assumptions about the concepts and without the need for example images for the concept. Hence, we focus on the description of an unsupervised discovered set of CAVs. As the discovered CAVs are given as vectors in the feature space, the encoded concepts need to be described for humans. 
A common way is to show images of a given dataset, which are most similar to the respective CAV in the hidden representation. However, this introduces the need for interpretation to derive a compact and communicable meaning from the given images.
 
To avoid the need for human interpretation, we propose to determine a ranking of textual descriptions for each concept. Depending on the CAV, the textual descriptions to be ranked, and the fine granularity of the text embedding, the highest ranked descriptions can be highly redundant. Therefore, we further derive a single common description based on the $k$ highest ranked descriptions. Depending on the ranking, this common description can differ from the highest ranked description.

We build up on existing approaches to describe the information filtered by individual neurons in a textual way, for example, \cite{oikarinenCLIPDissectAutomaticDescription2023}. In this approach a neuron is described by generating a textual description for the relevant images of a neuron for which the neuron has the highest activation. The textual description is chosen as the best fitting one out of multiple candidates. In contrast to individual neurons, a major advantage of CAVs is that they represent vectors in the feature space and not only individual scalar neuron outputs. The total number of CAVs is usually significantly lower than the number of neurons in the corresponding layer.

The textual descriptions of the individual neurons in  \cite{oikarinenCLIPDissectAutomaticDescription2023} are based on the full images that are relevant for the considered neuron.
However, when the variety of images in a data set is not large enough, it is often not possible to separate highly correlated concepts, especially concepts of different degrees of abstraction, purely based on the full images.
One example of the issue of highly correlated concepts are the concepts \textit{insects} and \textit{verde}, see \autoref{fig:concept_issues_left}. 
An example of concepts of different degrees of abstraction are the concepts \textit{underwater} and \textit{blue}, as in many cases \textit{underwater} is a specification of \textit{blue}, see \autoref{fig:concept_issues_right}.
To address this limitation, we propose to use receptive fields instead of the full images for the generation of the textual descriptions.
By replacing the full images with receptive fields, we can focus on the parts of the images, where an evaluated concept is most present.
This reduces the noise that can affect the textual description of the concept.

In summary, the interpretation process of a neural network by ranking textual descriptions of human understandable concepts is represented by CAVs. Further, we derive a single common textual description to decrease the redundancy.
Our main contributions are:

\begin{itemize}
    \item We enhance the automatic concept discovery in a trained model by interpreting the visual CAVs with textual descriptions.
    \item We derive a common concept description from the top-$k$ computed textual descriptions to reduce redundancy.
    \item We propose using receptive fields to derive the textual descriptions and introduce concept scores to measure the relevance of the receptive fields. By that the textual descriptions focus on the relevant parts of the images, e.g. only the parts of the image seen by the model up to that layer.
\end{itemize}

%% file: sec/related_work.tex
\section{Related Work}
\noindent\textbf{Concepts.}
The idea that certain directions in a model's latent representation align with human-understandable concepts was initially proposed by Kim et al. \cite{kimInterpretabilityFeatureAttribution2018}. They propose to learn a hyperplane in the activation space of a neural network layer that separates images, which include the concept, from other images. The normal of the hyperplane in the direction of the images encoding the concept is the Concept Activation Vector (CAV). Since then, a lot of effort was put into the automatic discovery of such concepts activation vectors \cite{fel2023craft,ghorbani2019towards,oikarinenLabelFreeConceptBottleneck2023,yeh2020completeness,zhang2021invertible}. Interesting for our work is the novel concept discovery algorithm proposed by Yeh et al. \cite{yeh2020completeness}, which combines interpretability with a new notion of \textit{completeness} which measures how sufficient a set of CAVs is for the explanation of a model's prediction behavior. They also introduce a method to rank the found CAVs by importance called ConceptSHAP which adapts Shapley values \cite{Roth_1988}. Shapley values assign importance to a feature by calculating its average contribution in all possible combinations. One drawback of approaches for automatic CAV discovery is that they rely on images as references for the explanation of a CAV.

\noindent\\\textbf{Network Dissection.} The idea of dissecting a network is to inspect the function of individual neurons in the network to get insights into the model. 
The first work about network dissection provided a method to quantify the interpretability of latent representations by comparing neuron activations with segmentation masks from a concept dataset \cite{bauNetworkDissectionQuantifying2017}. 
This approach aligns individual neuron activations of a model with specific visual concepts given by the segmentation masks. 
One major limitation of this approach is, that the masks needed to be annotated by humans. Based on this, a segmentation model was proposed in \cite{bauUnderstandingRoleIndividual2020} to annotate the masks for each concept. 
MILAN \cite{hernandezNaturalLanguageDescriptions2022} extends the labeling of neurons to open-ended natural language descriptions: 
This approach generates descriptions of neurons by finding language strings that maximize the mutual information of the image regions where the neuron is active. To generate the language description, an image-to-text model is required, trained on a labeled data set. To avoid the need for labeled data, CLIP-Dissect \cite{oikarinenCLIPDissectAutomaticDescription2023} leverages the multimodal training of CLIP \cite{radfordLearningTransferableVisual2021}, a method that embeds image and text data to a joint feature space.

\noindent\\\textbf{Joint Text-Image Embeddings.}
In recent years, there have been significant advancements in learning joint text-image embeddings \cite{jia2021scaling,li2019visualbert,radfordLearningTransferableVisual2021,zhai2023sigmoid}. 
Text-image embeddings can be utilized to perform various tasks, such as zero-shot classification. 
Contrastive learning based approaches, such as CLIP \cite{radfordLearningTransferableVisual2021}, are trained to maximize the similarity between positive examples (e.g., images and matching image captions) and to minimize the similarity to negative examples (e.g., non-matching image-caption pairs). Approaches such as CLIP have shown good zero-shot image classification performance on multiple data sets by evaluating the similarity between the feature embeddings of the class labels and the images.

\noindent\\\textbf{Post-Hoc Concept-Bottleneck Models.}
An alternative approach to gen\-er\-at\-ing post-hoc concept explanations is to first create a set of known CAVs and then find the subset of those CAVs that yield the best performance for a given model \cite{moayeriTextToConceptBackCrossModel,yuksekgonulPosthocConceptBottleneck2023}. Those approaches assume to have CAVs for all important concepts and then select the CAVs that can describe the essence of what was learned by the model. In our approach, the set of CAVs is discovered automatically by inspecting the model in more detail like in \cite{yeh2020completeness}, and then designated by textual descriptions.

%% file: sec/method.tex
\section{Method}
\begin{figure*}[ht!]
    \centering
    \includegraphics[width=\textwidth]{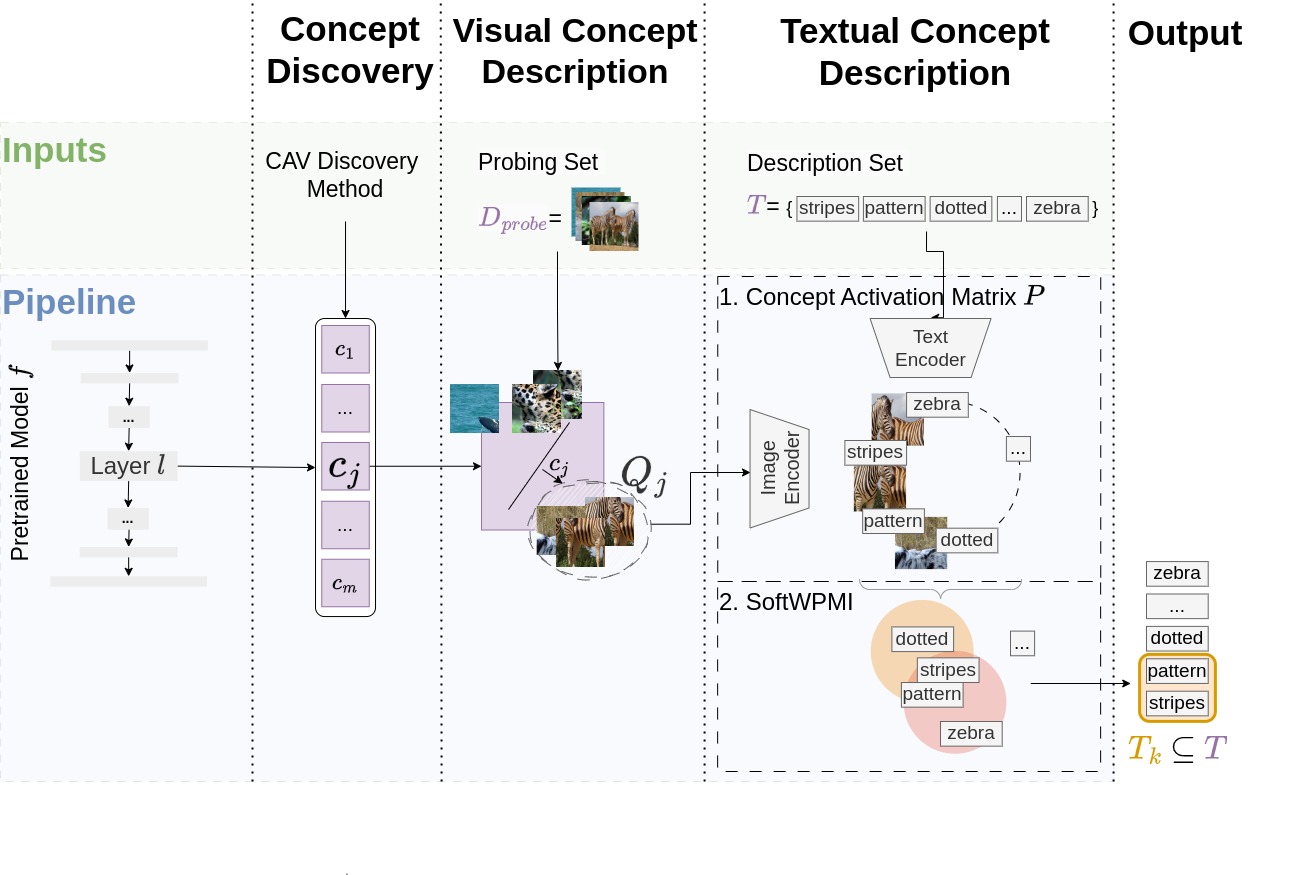}
    \caption{Overview of our approach to describe the layer $l$ of a pretrained model $f$.
    The \textit{inputs} are a concept discovery method, a probing set $\color{input}D_{probe}$, and a set of textual descriptions $\color{input}T$.
    We apply \textit{concept discovery} methods to find a set of CAVs, generate a set of \textit{visual concept descriptions} $Q_j$ for each CAV $c_j$, then \textit{textual concept descriptions} and finally \textit{output} the top-$k$ descriptions $\color{output}T_k \subseteq\textcolor{input}T$.
    }
    \label{fig:MethodConceptTextMatching}
\end{figure*}

We propose a method that derives textual descriptions for the concepts a neural network utilizes to solve an image classification task. The method consists of three steps, and each step represents a different level of concept description for a given neural network:
\begin{enumerate}
    \item The \textbf{discovery of concepts} by concept activation vectors (CAVs), represented as directions in the feature space, 
    \item the \textbf{visual description} of the concepts (encoded by the CAVs) with representative images,
    \item and the \textbf{textual description} of the concepts with words. 
\end{enumerate}
The steps are visualized in Figure \ref{fig:MethodConceptTextMatching}. In the following, the inputs, the three steps of the method, and the computed outputs are introduced in more detail.

\begin{figure*}[ht!]
    \centering
    \includegraphics[width=\textwidth]{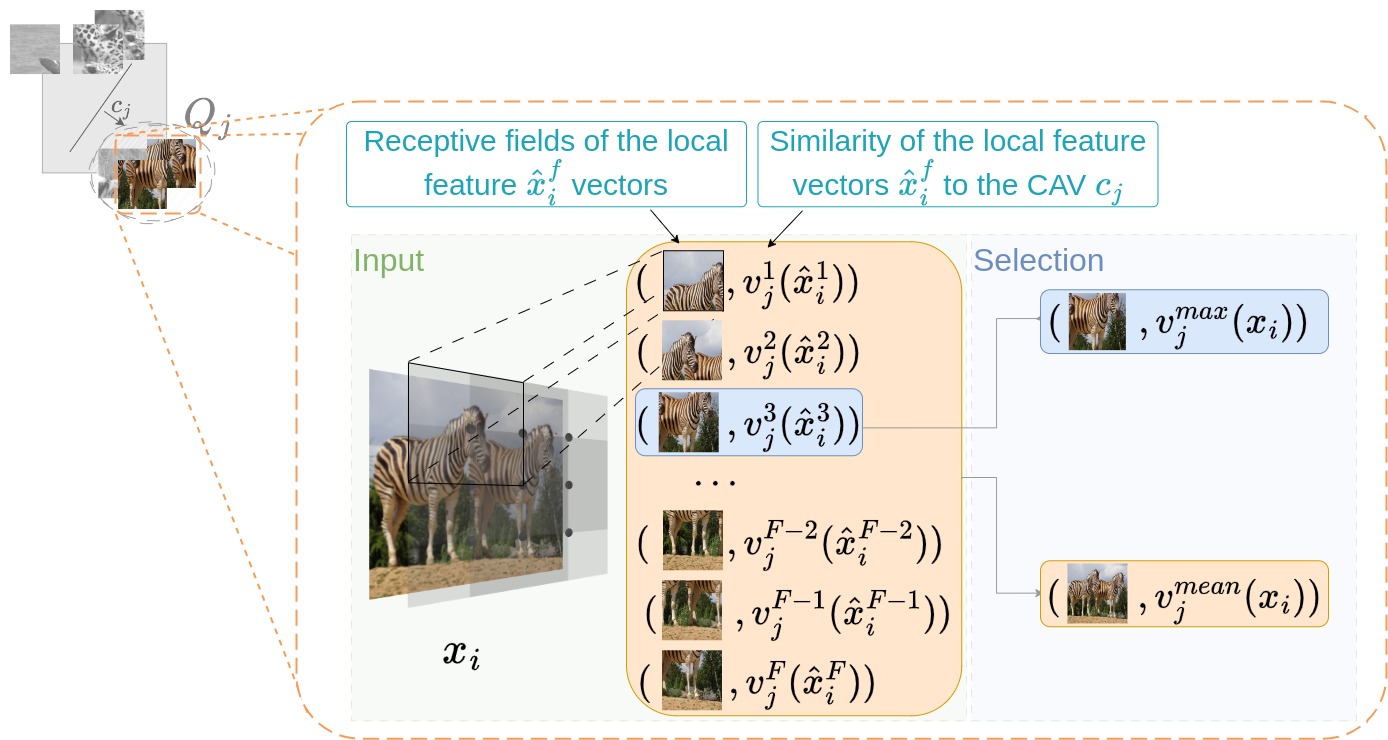}
    \caption{Selection of the visual representations for a given CAV $c_j$, compare with \autoref{fig:MethodConceptTextMatching} column \textit{Visual Concept Description}. 
    The vector $(v^1_j(\hat{x}^1_i), \dots, v^F_j(\hat{x}^F_i))$ represents the concept scores between each receptive field of $x_i$ and the CAV $c_j$. 
    While \cite{oikarinenCLIPDissectAutomaticDescription2023} select full images based on the mean score of all receptive fields, we also consider the receptive field with the highest concept score.
    Thus, we improve the visual input of the joint vision-text embedding by cropping $x_i$ to the respective receptive field.
    This creates a more truthful and more detailed representation of the concepts learned in the hidden space.}
    \label{fig:Selection}
\end{figure*}

\noindent\textbf{Inputs.} The method is based on a neural network trained on an image classification task, $f$, that maps input images to a $K$-dimensional output vector representing class probabilities. 
For a given layer $l$, for which concepts shall be extracted from the network, the network is decomposed into two functions $h_l$ and $\phi_l$, such that $f=h_l\circ\phi_l$. Further, let  $\mathcal{D}_{probe} =  \{x_1, \dots, x_n\}$ be a probing set, i.e., a set of $n$ images that can be used for the visual description of the extracted CAVs. 
The textual descriptions of the concepts are based on a predefined and task dependent set of words $T$. For example, $T$ can contain describing attributes \cite{bauNetworkDissectionQuantifying2017}, or the top 20.000 words of the English language \cite{First20hoursGoogle10000englishThis}.

\noindent\\\textbf{Concept Discovery.} We describe the embedding of layer $l$ with Concept Activation Vectors (CAVs). A CAV is a vector that points in the direction of a concept learned by the model and is embedded in the feature space of the activations of layer $l$. The concepts learned at layer $l$ are then represented by a set of $m$ CAVs, $C_l=\{c_{1,l},\dots,c_{m,l}\}$. We drop the index $l$ in the following when considering only one specific layer. While the proposed method is independent of the underlying concept extraction approach, we follow the approach of Yeh et al. \cite{yeh2020completeness} to derive all concepts utilized for a given image classification task.

\noindent\\\textbf{Visual Concept Description.} For the visual description of a given CAV $c_j$, we follow the former work \cite{yeh2020completeness} to derive a set $Q_j\subset\mathcal{D}_\text{probe}$ of most relevant images from the probing set. This approach is illustrated in Figure \ref{fig:Selection} and will be described in the following. The relevance of an image $x_i\in\mathcal{D}_\text{probe}$ is determined based on the similarity between the CAV $c_j\in\mathbb{R}^k$ and its latent representation at layer $l$. In detail, consider the latent representation of an image $x_i$ at layer $l$, which is $$\phi_l(x_i)=:(\hat{x}^1_{i,l}, \dots, \hat{x}^F_{i,l})\in\mathbb{R}^{F\times k}.$$
The vectors $\hat{x}^1_{i,l}, \dots, \hat{x}^F_{i,l}$ are called \textit{local feature vectors} of $x_i$ and correspond to the activations of each channel of the convolutional neural network after layer $l$.
We will omit the index $l$ when the connection to the specific layer is clear. 

\noindent For each local feature vector $\hat x_i^f$ and each CAV $c_j$ a \textit{concept score}, which measures the similarity based on the scalar product, i.e.
$$v^f_j(\hat{x}^f_i):=\hat{x}^{f_i^T} c_j.$$
This leads to a vector $v_j(x_i)\in\mathbb{R}^F$ of $F$ concept scores, 
\begin{equation}
v_j(x_i)=\left(v^1_j(\hat{x}^1_i), \dots, v^F_j(\hat{x}^F_i)\right)\in \mathbb{R}^F.
\label{eq:score_vector}
\end{equation}
Following \cite{yeh2020completeness}, a larger concept score means a higher similarity of the cor\-re\-spond\-ing receptive field of $\hat{x}^f_i$ to the concept encoded by the CAV $c_j$.

\noindent While $v_j(x_i)$ is a vector of similarities, the set of relevant images $Q_j$ is chosen based on scalar values because they can be ordered. 
Former works such as \cite{oikarinenCLIPDissectAutomaticDescription2023} select full images of $\mathcal{D}_\text{probe}$ for the set $Q_j$. To achieve this, they consider the average over the individual concept scores of the local feature vectors, 
\begin{align} \label{eq:concept_score_mean}
    v_j^{\text{mean}}(x_i)=\tfrac{1}{F}\sum^F_{f=1}{v^f_j(\hat x_i^f}) \in\mathbb{R}.
\end{align}
As we are more interested in the most representative part of an image for a concept, we consider the maximum concept score of all local feature vectors:
\begin{align} \label{eq:concept_score_max}
    v_j^\text{max}(x_i)=\max_{f\in\{1,..,F\}}{v^f_j(\hat x_i^f}) \in\mathbb{R}.
\end{align}

\noindent Based on these two metrics, we introduce three different strategies to derive a set of most relevant images $Q_j$ from $\mathcal{D}_\text{probe}$. Note that the subset $Q_j$ can either contain the full image $x_i$ or a receptive field associated with a local feature vector $\hat{x}^f_i$. We follow [22] and select the 100 most relevant images.
\begin{itemize}
    \item \MeanRF: Select the images with the highest $v_j^{\text{mean}}(x_i)$.
    \item \MaxRF: Consider those images with the highest $v_j^\text{max}(x_i)$ and choose the respective receptive fields where the maximum is reached.
    \item \MeanMaxRF: Select images like \MeanRF but choose the receptive field with highest concept score $v_j^f(\hat x^f_i)$.
\end{itemize}

We search for the parts of the images with the highest presence of the concept encoded by the CAV. With \MeanRF we select the full images with the highest overall presence of the concept. 
As a result, the textual descriptions are calculated based on the full images.  
However, often the model can only see parts of the images at the layer where the CAVs were found. 
Due to this, and the fact that concepts may be more present in single parts of an image, we apply strategies to find the relevant receptive fields.
Using \MaxRF we select the receptive field of each image with the highest concept score. 
We propose \MeanMaxRF to combine the advantages of both strategies. 
This means that we find the images where the concept is highly present in the full image and reduce the noise introduced by other concepts by selecting the respective receptive field with the highest concept score.

\noindent\textbf{Textual Concept Description. } To derive a textual description for the visual descriptions collected in $Q_j$, we utilize joint text-image embeddings and corresponding image and text encoders $E_\mathcal{I}$ and $E_\mathcal{T}$ which map from the space of images, $\mathcal{I}$, and the space of texts, $\mathcal{T}$, respectively, to a joint feature space. This is, for example, provided by the CLIP model \cite{radfordLearningTransferableVisual2021}. We compute a similarity matrix $P$ based on the cosine similarity of the text and image embeddings of the textual descriptions set $T=\{t_1,\dots,t_s\}$ and images in $Q_j$, 
$$P_{ij}=\tfrac{E_\mathcal{I}(x_i)^T E_\mathcal{T}(t_j)}{\Vert  E_\mathcal{I}(x_i)\Vert_2 \Vert  E_\mathcal{T}(t_j)\Vert_2}~.$$

\noindent Intuitively, we want to find the textual descriptions that have a high similarity to all images in $Q_j$.
To do this, we utilize the \textit{Soft Weighted Pointwise Mutual Information} (SoftWPMI) \cite{oikarinenCLIPDissectAutomaticDescription2023}, which indicates how well a word describes the mutual information of the representative images.
SoftWPMI requires a weighting of the images in $Q_j$, which is determined by the concept scores. In particular, this vector $q_j$ is calculated depending on the strategy to derive the set of most relevant images $Q_j$:
\begin{equation} \label{eq:weighting_strategy}
    q_j = 
    \begin{cases} 
    (v_j^{mean}(x_i))_{x_i\in Q_j} & \text{if } \text{\MeanRF} \\
    (v_j^{max}(x_i))_{x_i\in Q_j} & \text{if } \text{\MaxRF}, \text{\MeanMaxRF}
    \end{cases}
\end{equation}

Finally, we find the subset $T_k$ with the top-$k$ textual descriptions by:
\begin{equation} \label{eq:concept2word}
    T_k:=\argmax_{\hat T \subset T: |\hat T| = k} \sum_{t \in \hat T} \text{SoftWPMI}(t,q_j,P) 
\end{equation}
Note that, in practice, $\text{SoftWPMI}(t,q_j,P)$ is computed for each  $t\in T$ separately, and finally, we take the top-$k$ textual descriptions.
For the common textual description, we compute the weighted average of the top-$k$ descriptions in the feature space, with the weighting based on the SoftWMPI values. The common representation is then chosen as the textual description in $T$ that is closest to this weighted average. Please note that we set all negative SoftWMPI values in $\hat T$ to zero since we are only interested in positive similarities.

\noindent\textbf{Output.} The method returns the common description and the subset $T_k$ from the human understandable textual descriptions set $T$, which are most similar to the concept represented by the CAV $c_j$.

%% file: sec/experiments.tex
\section{Experiments}
Our experimental procedure consists of three stages. 
First, we utilize CAVs with known concept labels to show that our approach is capable to yield meaningful textual explanations of CAVs.
Second, we compare the different mappings \MeanRF, \MaxRF, \MeanMaxRF for the generation of the set of best fitting textual descriptions. 
And finally, we consider a more complex scenario and explain a set of CAVs extracted from a model where we have no prior knowledge about the underlying concepts.

\begin{table}[ht]
    \centering
    \caption{Each row shows the Top-5 textual descriptions of a CAV computed with the proposed approach (ranked from left to right) and the derived common concept description. Each CAV is supposed to represent one class of the CIFAR10 dataset \cite{krizhevsky2009learning}. Imagenet is utilized \cite{imagenet} as $\mathcal{D}_\text{probe}$ and google20k as the set of textual descriptions, $T$.}
    \label{tab:explicitly_class_concepts_cifar}
    \begin{adjustbox}{width=\columnwidth}
        \begin{tabular}{@{}l|c|ccccc@{}}
        \hline
            \toprule
            CAV-Label &  Common Description & 1 & 2 & 3 & 4 & 5 \\
            \midrule
            airplane & aircraft & aircraft & aviation & plane & airplanes & planes \\
            automobile & vehicle & vehicle & vehicles & car & ambulance & automobile \\
            bird & bird & avian & bird & birding & birds & juvenile \\
            cat & cat & cat & kitts & kitty & kitten & katz \\
            deer & deer & grazing & gnu & deer & female & wildlife \\
            dog & dog & puppy & dog & canine & pundit & dug \\
            frog & mating & mating & meal & head & emerging & frog \\
            horse & horse & equine & horseback & horse & horses & equestrian \\
            ship & sailing & sailing & yacht & sail & yachts & sailors \\
            truck & trucks & truck & trailer & trucks & trailers & movers \\
            \bottomrule
        \end{tabular}
    \end{adjustbox}
\end{table}

\subsection{Explaining a Set of CAVs with Known Concept Labels}
To be able to validate general idea of our approach, we follow Kim et al. \cite{kimInterpretabilityFeatureAttribution2018} and design a set of CAVs where each CAV describes one class of a given data set. We achieve this by generating a set of CAVs after the last convolutional layer of a model and set the number of CAVs equal to the number of classes.
It is important to note that the suggested strategy is closely related to the performance of the CLIP model. Hence, a bad classification performance of CLIP directly affects our approach in a negative way.

\noindent\textbf{Setup.} 
To make sure that the CLIP model itself performs well in this validation example, we use the datasets CIFAR10 \cite{krizhevsky2009learning} and MNIST \cite{deng2012mnist} which have a zero-shot performance of 96.2\% and 87.2\%, with the vision encoder CLIP-ViT L/14 from CLIP \cite{radfordLearningTransferableVisual2021}.
For CIFAR10 we adapted a pre-trained ResNet50 \cite{he2016deep} and finetuned it.
The finetuned ResNet50 reaches an accuracy of 0.94.
For MNIST we finetuned a simple ConvNet with 3 layers reaching an accuracy of 0.98.
In this experiments we explain the the embedding after the last convolutional layer of the models (ResNet50 and ConvNet).
For the set of textual descriptions we use google20k \cite{First20hoursGoogle10000englishThis}. Details to the MNIST experiments can be found in the appendix.

\noindent\textbf{Results.} 
The results of this experiment for CIFAR10 are displayed in \autoref{tab:explicitly_class_concepts_cifar} (The table for MNIST can be found in the Appendix). 
The top-5 words, as well as the concept closest to the centroid for each class, are shown. Our approach is able to match each CAV which encodes a class as concept with fitting textual descriptions from the 20.000 textual suggestions given. 
The exception is the CAV encoding \textit{Frog}.
For MNIST our approach finds fitting textual descriptions for all classes except the CAV encoding ``one'' which is described by \textit{makefile}.

\begin{figure*}[!htb]
    \centering
    \begin{subfigure}[t]{0.49\textwidth}
        \includegraphics[width=\linewidth]{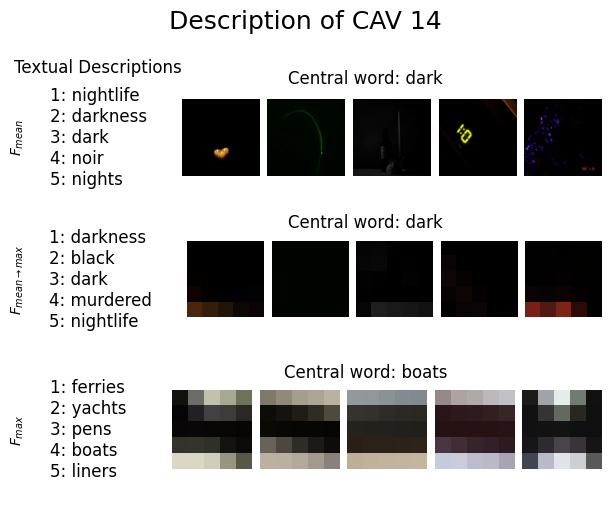}
        \caption{Most influencial CAV for the class ``cat''}
        \label{fig:dcatsvsdogs_left}
    \end{subfigure}
    \hfill
    \begin{subfigure}[t]{0.49\textwidth}
        \centering
        \includegraphics[width=\linewidth]{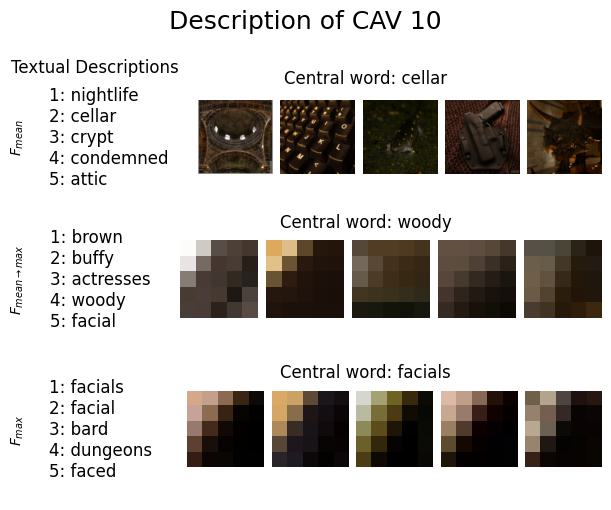}
        \caption{Second most influencial CAV for the class ``cat''}
        \label{fig:dcatsvsdogs_right}
    \end{subfigure}
    \vspace{-0.2cm}
    \caption{Comparison of the approaches to generate textual descriptions. Shown are the two most influential CAVs for the class ``cat'' after the first residual block of a ConvMixer \cite{trockman2023patches}. The model was trained on \textit{dark} cats and \textit{light} dogs, a subset of the Cats vs. Dogs dataset \cite{dogs-vs-cats}. The first approach uses the images with the highest mean activation for the CAV, the second takes the highest receptive fields of the images with the highest mean activation and the third takes the most activated receptive fields of all receptive fields over the whole probing data set. The probing dataset is the validation set from ImageNet \cite{imagenet} and the concept set is google20k \cite{First20hoursGoogle10000englishThis}} \label{fig:DCatsVsDogs_Discovery}
\end{figure*}

\subsection{Concept Discovery and Description}
Compared to the class-wise concepts in the previous sections, automatically discovered CAVs usually describe more abstract concepts as colors and shapes. 
We utilize the approach of \cite{yeh2020completeness} to discover a set of CAVs automatically. The final set of CAVs is selected based on a hyper parameter search and the test accuracy of the classification task.
The hyper parameter search includes the number of concepts,the threshold value $\beta$, and scalars $\lambda_1>0$ and $\lambda_2>0$. The parameters $\lambda_1$ and $\lambda_2$ are needed for the utilized concept discovery approach of \cite{yeh2020completeness}. They weight the similarity between the concepts and their most relevant images ($\lambda_1$) and the pairwise dissimilarity between the concepts ($\lambda_2$).
Further, we calculated for each class the ConceptSHAP and explanation quality following \cite{yeh2020completeness}.
The ConceptSHAP gives us an importance value for each CAV with respect to the class.
The explanation quality serves as a measure how well a class is described by the set of CAVs discovered. In the following, we first compare the different approaches to select the relevant images, i.e., the receptive field-based approaches and the full image approaches.

\subsubsection{Evaluation of Image Set Selection}
We consider concepts extracted from early layers, where concepts are assumed to be more abstract than in later layers. With this we can also evaluate the effect of \MaxRF and \MeanMaxRF on highly correlated concepts and concepts of different degree of abstraction.
We further introduce the abstract concept \textit{dark} into the model by performing a classification of cat and dog images, where the training samples consist of dark cats and the bright dogs.
We expect the trained model to mainly rely on those features due to the simplicity bias of neural networks \cite{shah2020pitfalls}.

\noindent\textbf{Setup.} 
We trained our model on a modified Cats vs. Dogs (CvD) dataset \cite{dogs-vs-cats}.
The Cats vs Dogs dataset was developed by Kaggle \cite{dogs-vs-cats} and, following \cite{lakkaraju2016discovering,kim2019learning}, we split it by color, such that it consists of \textit{dark} cats and \textit{light} dogs.
We call this dataset Dark Cats vs. Dogs (DCvD).
In the following we refer to the original and the modified dataset as unbiased and biased dataset.
Since all cats are \textit{dark} and all dogs are \textit{light} we make the assumption, that the color is a relevant concept for models trained on this dataset.
To validate this we train a ConvMixer \cite{trockman2023patches} with a depth of seven. 
The ConvMixer reaches an accurcay of ~0.93 on the biased data and only an accuracy of of ~0.69 on the unbiased data (details in the appendix).
This difference in accuracy indicates that the model learned to associate the color \textit{black} with cats. 
We extract the set of CAVs after the first residual block of the model.
The derived set consists of 20 CAVs and the classification based on the active and inactive CAVs yields an accuracy of ~0.96 on the biased data. 
The hyper parameters used to learn the set are $\lambda_1=0.2$, $\lambda_2=0.2$ and $\beta=0.18$. 
After we filter the CAVs where the dot product is over 0.95 we are left with 15 relevant CAVs.
As the set of textual descriptions, google20k is used.

\noindent\textbf{Results.} 
\autoref{fig:DCatsVsDogs_Discovery} shows the two most important CAV from left to right for the class cat. The CAVs are selected by the ConceptSHAP values.
For each CAV we display the three approaches to select relevant images based on the concept scores. 
For each approach the textual descriptions and the top five images from the set of most relevant images are shown. 
It can be seen for CAV 14 that the approach \MeanRF returns as highest textual description \textit{nightlife} and \MaxRF returns \textit{ferries} (See \autoref{fig:dcatsvsdogs_left}). 
Only \MeanMaxRF returns a fitting highest textual description with \textit{darkness}.
Looking at the other descriptions we see that \MeanRF also yields similar textual descriptions in the top 5 descriptions.
This results in the central word of \MeanRF and \MeanMaxRF, matching our expectations.

\noindent For the CAV 10 we can see that all approaches return different textual descriptions (See \autoref{fig:dcatsvsdogs_right}). 
\MeanRF returns \textit{nightlife} and \MaxRF returns \textit{facials} which are both complex concepts. The approaches recognize different concepts which are relevant for the images. This is neither good nor bad.
Only \MeanMaxRF returns a simple concept with \textit{brown}.

\subsubsection{Animals with Attributes}

\begin{figure*}[!htb]
    \centering
    \begin{subfigure}[t]{0.49\textwidth}
        \includegraphics[width=\linewidth]{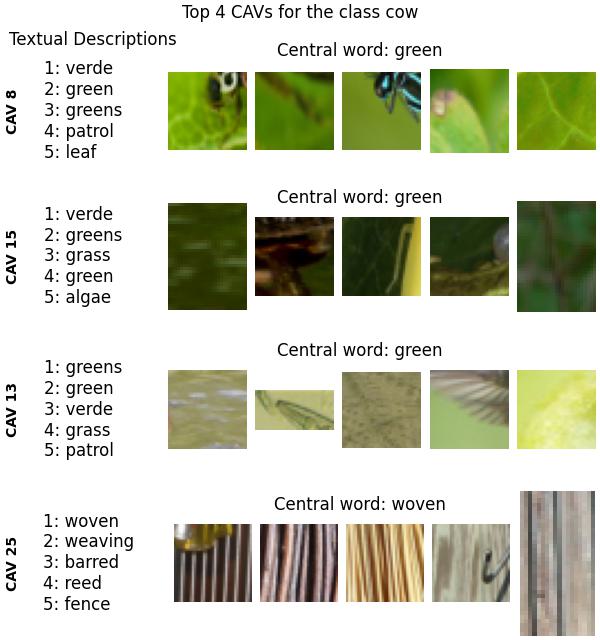}
        \caption{Description of the best represented class: ``cow''}
        \label{fig:awa_left}
    \end{subfigure}
    \hfill
    \begin{subfigure}[t]{0.49\textwidth}
        \centering
        \includegraphics[width=\linewidth]{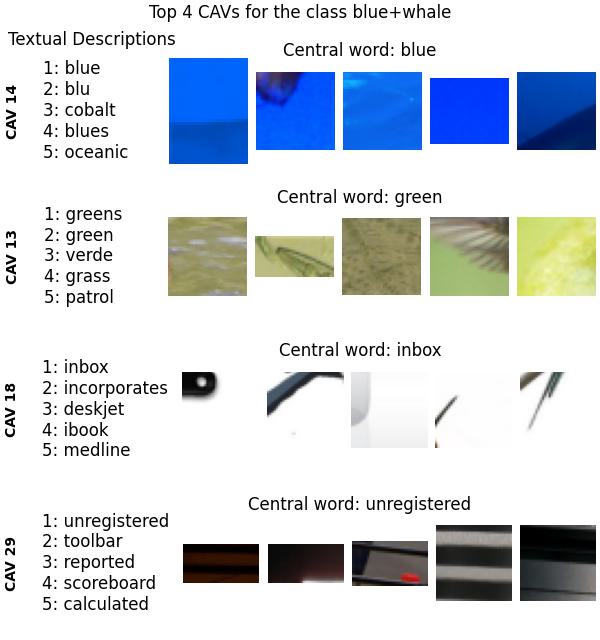}
        \caption{
        Description of the worst represented class: `blue whale''
        }
        \label{fig:awa_right}
    \end{subfigure}
    \vspace{-0.2cm}
    \caption{
    For each class the textual descriptions and the most activated receptive fields of the CAVs with the strongest influence are shown. The image set was selected by \MeanMaxRF. The set of CAVs describes the hidden representation after the first residual bock of a ResNet50 finetuned on AwA2. The probing dataset is the validation set from ImageNet and the concept set is google20k.
    }
    \label{fig:concept_discovery_figure}
\end{figure*}

The objective of this experiment is to explore the performance of our approach for scenarios with increased complexity and to show its potential.
The experiment is based on the Animals with Attributes2 dataset \cite{xian2018zero}, which contains 37322 images from 50 different animals.

\noindent\textbf{Setup.} 
We finetuned a ResNet50 on the dataset AwA2 \cite{xian2018zero} that reaches a test accuracy of ~0.9. 
The concept discovery method found a set of 30 CAVs after the first residual block. 
The found set of CAVs achieves an accuracy of ~0.87 with the hyper parameter $\lambda_1=3.1$, $\lambda_2=3.1$ and $\beta=0.02$. 
After filtering all duplicates 15 CAVs are left, describing the concepts learned by the first residual block.

\noindent\textbf{Results.} The results of this experiment can be seen in \autoref{fig:concept_discovery_figure}. 
Here, \autoref{fig:awa_left} shows the class which is best described by the set of CAVs and
\autoref{fig:awa_right} shows the class which is worst described by the set of CAVs.
Further, for each class the most influential CAVs ranked by ConceptSHAP are displayed. 
The descriptions are generated with the \texttt{\MeanMaxRF} approach. 
It can be observed that the model strongly connects the concept \textit{green} with the class ``cow'' (See \autoref{fig:awa_left}).
The class ``blue whale'' is connected to the concept \textit{blue} (See \autoref{fig:awa_right}). 
When inspecting the descriptions of the CAVs 18 and 29 a mismatch becomes apparent. 
The descriptions for those CAVs seem to be hardly related and are not matching to the receptive fields.

%% file: sec/discussion.tex
\section{Discussion} 
The experiments on the sets of CAVs with the known concept label show that the approach is capable of matching CAVs with the corresponding textual descriptions from a large set of general descriptions. This underlines that our approach is in general capable of identifying joint textual descriptions, even though the performance highly depends on the quality of the utilized joint text-image features space. For the experiment on CIFAR10, one can further see the redundancy in the best-fitting descriptions which is successfully removed by selecting a common concept description (\autoref{tab:explicitly_class_concepts_cifar}). Further, one can see the approach's capabilities to detect biases in the training and/or probing images, e.g., the top five descriptions of the class \textit{ship} are all related to sailing.

For the different approaches to select representative images for given CAVs, the ones using receptive fields help to correctly describe more abstract concepts that especially occur in earlier layers of a neural network (\autoref{fig:DCatsVsDogs_Discovery}).
Interestingly, 15 CAVs are detected as relevant, which is more than to separate the concepts of dark and bright. This can be explained by the fact, that dark and bright colors can also occur in the backgrounds of the images and hence the distinction purely based on color concepts is not feasible. However, the relevance of the \textit{dark} concept shows that it is highly relevant to classify cats. The increased focus on abstract concepts when utilizing the receptive fields can also be explained by the nature of the CLIP model. CLIP was trained on images and corresponding captions, where specific colors (e.g., \textit{green}) might be less relevant than the overall image description (e.g., \textit{insect}).
In \autoref{fig:awa_right}, the CAVs 18 and 29, which are relevant for the class ``blue whale'', are examples where the approaches fail to generate matching textual descriptions. This can be attributed to limitations in the utilized CLIP model.
For example, CAV 18 seems to show the concept \textit{white} but the textual descriptions are \textit{inbox}, \textit{incorporate}, \dots.
This could be improved by applying a more fine-grained selection of the inputs for the joint text-image model or by utilizing other text-image feature spaces.

%% file: sec/conclusion.tex
\section{Conclusion}
In this work, we proposed an approach to assist the interpretation of CAVs by suggesting textual descriptions and selecting common words for the individual CAVs.
To improve the textual descriptions of CAVs found for earlier layers, we consider that for earlier layers of a model, the CAVs do not know the whole input and propose to use receptive fields for the generation of the textual descriptions.
Through experiments on sets of CAVs where the underlying concepts are known, we showed that our method is capable of yielding meaningful descriptions for CAVs and that the usage of receptive fields improves the explanation quality for earlier layers. 
While this research already offers insights into the concept discovery process, further works on the computation of meaningful concepts as well as an exploration of other image-to-text projections are planned. The evaluation of the found textual descriptions regarding human understanding is also a topic for further research.
To better understand the behaviour of the model, it would be interesting to extend the results of concept discovery methods with mismatched data. 
For the description of specific concepts, further insights into the capabilities of joint text-image feature spaces and the needed characteristics of probing sets are interesting for us, as well as the consideration of explicitly fine-tuning text-image embeddings to basic concepts.

%% file: sec/acknowledgements.tex
\subsubsection{Acknowledgements} We thank Niklas Penzel for preparing the Dark Cats vs. Dogs (DCvD) dataset and training the corresponding model.

%% file: sec/appendix.tex
%







\title{Exploiting Text-Image Latent Spaces for the Description of Visual Concepts}


\clearpage

\begin{center}
\textbf{\large Appendix: \mytitle}
\end{center}
\setcounter{equation}{0}
\setcounter{figure}{0}
\setcounter{table}{0}

\section{Additional Examples}
\subsection{Explaining a Set of CAVs with Known Concept Labels}
\begin{table}[ht]
    \centering
    \caption{Top-5 closest descriptions are shown from left to right and joint concept description for each CAV. The CAVs encode the classes from MNIST}
    \label{tab:explicitly_class_concepts}
    \begin{adjustbox}{width=\columnwidth}
        \begin{tabular}{@{}l|c|ccccc@{}}
        \hline
            \toprule
            CAV-Label & Common Description & 1 & 2 & 3 & 4 & 5 \\
            \midrule
            zero & circular & circular & ring & rings & circle & oval \\
            one & domains & makefile & hostname & indices & authored & deprecated \\
            two & twenty & twentieth & two & twenty & second & twelve \\
            three & three & three & tres & thirds & iii & third \\
            four & four & four & fourth & fourteen & forty & quad \\
            five & five & five & fifth & sixth & fifteen & fifty \\ 
            six & sixty & sixty & om & viii & lev & horns \\
            seven & seven & seven & seventh & vii & seventy & hebrew \\
            eight & eight & eight & eighty & infinite & nine & infinity \\
            nine & nine & nine & eight & ninth & ninety & eighty \\
            \bottomrule
        \end{tabular}
    \end{adjustbox}
\end{table}

In \autoref{tab:explicitly_class_concepts} we present the results of our approach for MNIST \cite{deng2012mnist}. It can be seen, that the approach yields matching descriptions for all CAVs except \textit{one}.

\subsection{Evaluation of Image Set Selection}
\begin{figure*}[!htb]
    \centering
    \begin{subfigure}[t]{0.49\textwidth}
        \includegraphics[width=\linewidth]{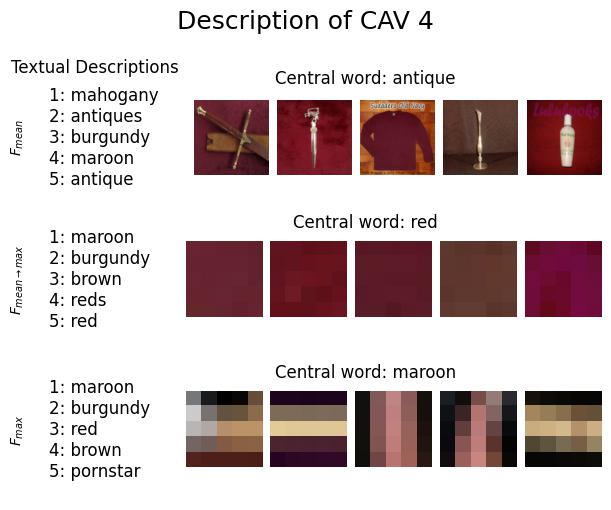}
    \end{subfigure}
    \hfill
    \begin{subfigure}[t]{0.49\textwidth}
        \centering
        \includegraphics[width=\linewidth]{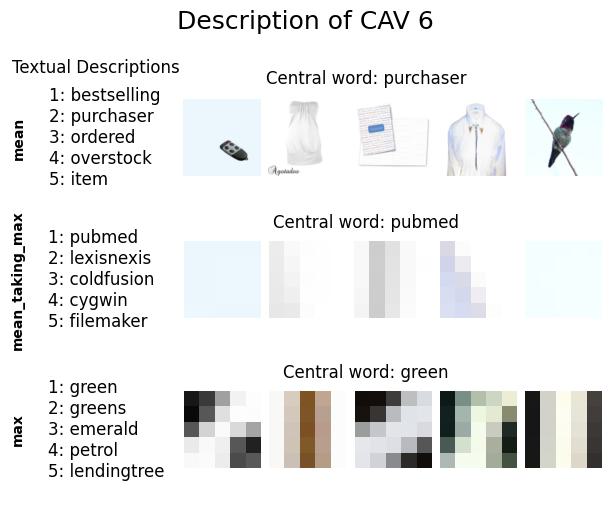}
    \end{subfigure}
    \hfill
    \begin{subfigure}[t]{0.49\textwidth}
        \centering
        \includegraphics[width=\linewidth]{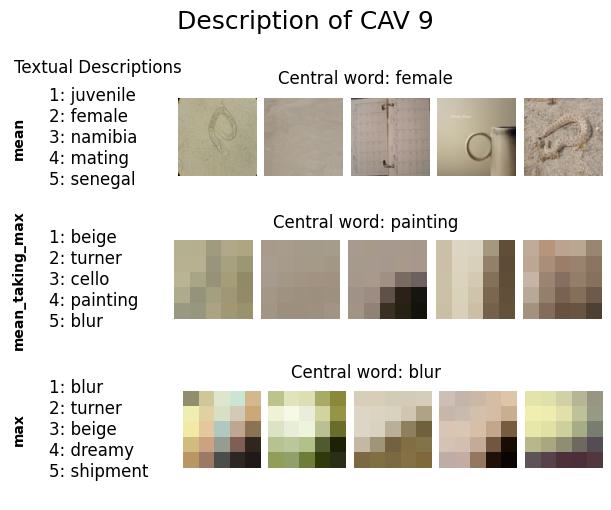}
    \end{subfigure}
    \hfill
    \begin{subfigure}[t]{0.49\textwidth}
        \centering
        \includegraphics[width=\linewidth]{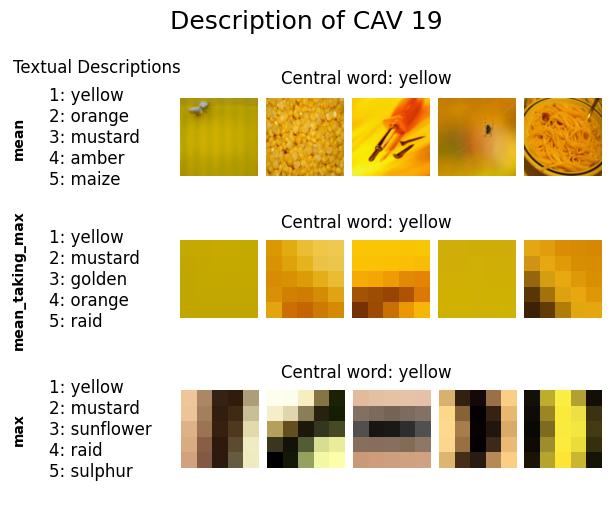}
    \end{subfigure}
    \vspace{-0.2cm}
    \caption{Comparison of the approaches to generate textual descriptions. Shown are influential CAVs for the class ``dog'' after the first residual block of a ConvMixer \cite{trockman2023patches}. The probing dataset is the validation set from ImageNet \cite{imagenet} and the concept set is google20k \cite{First20hoursGoogle10000englishThis}} \label{fig:DCatsVsDogs_Discovery}
\end{figure*}
In \autoref{fig:DCatsVsDogs_Discovery} we present additional CAVs which are important for the class ``dog''. For each CAV we show the different approaches to generation of the set of best fitting textual descriptions.

\subsection{Animals with Attributes}
\begin{figure*}[!htb]
    \centering
    \begin{subfigure}[t]{0.49\textwidth}
        \includegraphics[width=\linewidth]{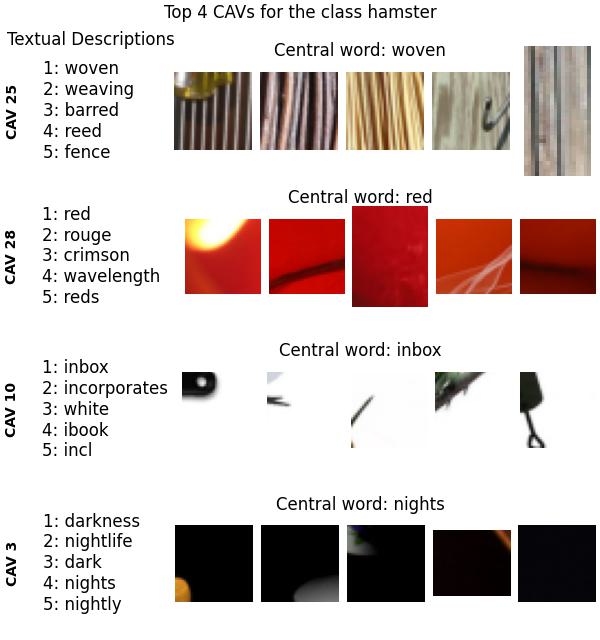}
    \end{subfigure}
    \hfill
    \begin{subfigure}[t]{0.49\textwidth}
        \centering
        \includegraphics[width=\linewidth]{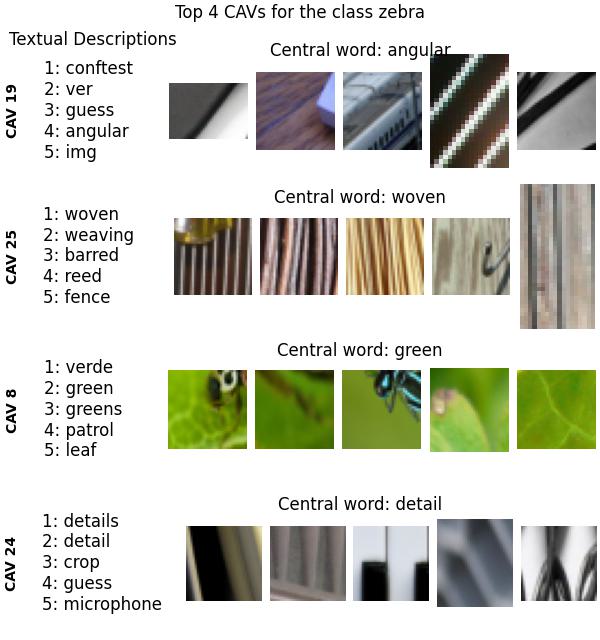}
    \end{subfigure}
    \hfill
    \begin{subfigure}[t]{0.49\textwidth}
        \centering
        \includegraphics[width=\linewidth]{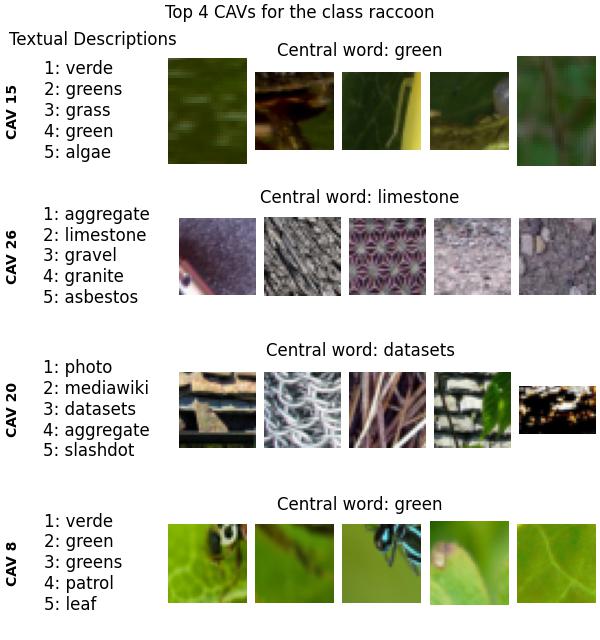}
    \end{subfigure}
    \hfill
    \begin{subfigure}[t]{0.49\textwidth}
        \centering
        \includegraphics[width=\linewidth]{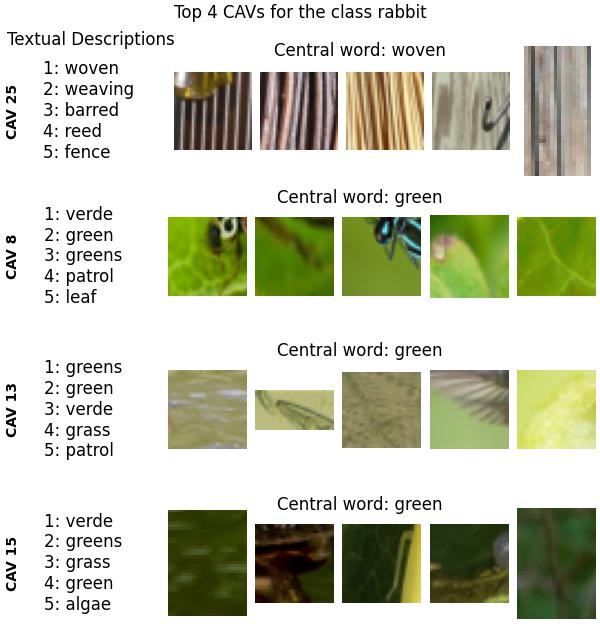}
    \end{subfigure}
    \vspace{-0.2cm}
    \caption{Description of the classes ``hamster'', ``zebra'', ``raccoon'' and ``rabbit'' according to a set of CAVs. For each class, the textual descriptions and the most activated receptive fields of the CAVs with the strongest influence are shown. The image set was selected by \MeanMaxRF. The set of CAVs describes the hidden representation after the first residual block of a ResNet50 \cite{krizhevsky2009learning} finetuned on AwA2 \cite{xian2018zero}. The probing dataset is the validation set from ImageNet \cite{imagenet} and the concept set is google20k\cite{First20hoursGoogle10000englishThis}} \label{fig:awa}
\end{figure*}
In \autoref{fig:awa} we show more textual descriptions for different classes from the AwA dataset \cite{xian2018zero}.

\section{Models}
\noindent\textbf{ConvNet.}
For the MNIST dataset, we trained a model with the following specifications:
The model included three convolutional layers with inner channel sizes of 32, 64, and 128. We used cross-entropy as the loss function and Adam as the optimizer, with a learning rate of 0.001 and a weight decay of 0.0005.
During training, we implemented early stopping. We monitored validation loss with a patience of 10 epochs and training accuracy with a patience of 15 epochs. The maximum number of epochs was set to 1000.

\noindent\textbf{ConvMixer.}
For the Dark Cats vs Dogs dataset \cite{dogs-vs-cats}, we trained a ConvMixer \cite{trockman2023patches} model with the following specifications:
The model had a dimension of 256, a depth of 8, a kernel size of 7, and a patch size of 5. We used cross-entropy as the loss function and AdamW as the optimizer, with a learning rate of 0.001 and a weight decay of 0.0005.
We performed data transformations by resizing the images to 128x128 pixels, using TrivialAugment \cite{muller2021trivialaugment} for augmentation, and normalizing the images with mean values of 0.5, 0.5, 0.5, and standard deviation values of 0.5, 0.5, 0.5.
During training, we implemented early stopping. We monitored validation loss with a patience of 10 epochs and training accuracy with a patience of 15 epochs. The maximum number of epochs was set to 1000.

\noindent\textbf{Finetuned ResNet50s.}
For the CIFAR-10 \cite{he2016deep} and the Animals With Attributes dataset \cite{xian2018zero}, we finetuned a ResNet50 model \cite{krizhevsky2009learning} with the following specifications:
We used cross-entropy as the loss function and Adam as the optimizer, with a learning rate of 0.001 and a weight decay of 0.0005.
During training, we implemented early stopping. We monitored validation loss with a patience of 10 epochs and training accuracy with a patience of 15 epochs. The maximum number of epochs was set to 1000.